# Robotic design choice overview using Co-simulation

[1,2,*]Martin Peter Christiansen, [1]Peter Gorm Larsen, [1]Rasmus Nyholm Jørgensen

[1]Aarhus University, Department of Engineering, Finlandsgade 22, 8200 Aarhus N, Denmark
[2]Conpleks Innovation, Fælledvej 17, 7600 Struer, Denmark

**Abstract:** Rapid robotic system development sets a demand for multi-disciplinary methods and tools to explore and compare design alternatives. In this paper, we present collaborative modelling that combines discrete-event models of controller software with continuous-time models of physical robot components. The presented co-modelling method utilized VDM for discrete-event and 20-sim for continuous-time modelling. The collaborative modelling method is illustrate with a concrete example of collaborative model development of a mobile robot animal feeding system. Simulations are used to evaluate the robot model output response in relation to operational demands. The result of the simulations provides the developers with overview of the impacts of each solution instance in the chosen design space. Based on the solution overview the developers can select candidates that are deemed viable to be deployed and tested on an actual physical robot.

## Introduction

When developing automatic robotic systems generally the overall development goal is to enable the robot to perform the desired tasks based on the overall system demands [1]. Robotic system development utilizing modelling and simulation is an approach that gradually is adopted as an integral part of the process [2], [3], [4]. Modelling provides the developers with the capabilities to explore hardware and software solutions before developing the actual component. The modelling and simulation angle allows for the automatic evaluation of a much larger potential design space compared to a manual trial and error approach. The alternative development approach for robotic systems involves significant time spent on ad-hoc trial and error testing, to reach a usable system configuration of the physical system. In this case developers can end up spending valuable development-time on finding optimal solutions in areas, which in the final solution shows little impact on the desired outcome.

The prime challenge here is that many complementary disciplines are necessary to determine viable solutions i.e., electrical; mechanical; software; embedded systems and signal processing [5], [6]. Each discipline has different cultures, tools and methodologies, which can restrict the development of a cross-disciplinary project. In this paper, we propose a model-based simulation approach with collaborative modelling, enabling a combination of models from the different disciplines.

Collaborative simulations (co-simulations) allows developers to examine different aspects of the system to explore design alternatives. Co-simulations are based on the models the developers utilise to describe the different aspects of the robotic system. Co-modelling and co-simulation are performed using the Crescendo technology[1] produced in the European DESTECS FP7 project[2].
Design Space Exploration (DSE) is the analysis of different candidate solutions using co-simulation. The idea behind DSE is to estimate which candidates the developers can be deployed on the actual system to perform the desired result. The aim is to illustrate the exploration of a cross-disciplinary robotic design challenge using co-simulation. The robotic design challenge is based on a robotic feeding system for agricultural farming applications (see **Fejl! Henvisningskilde ikke fundet.**).

---

[1] See http://crescendotool.org
[2] See http://destecs.org/

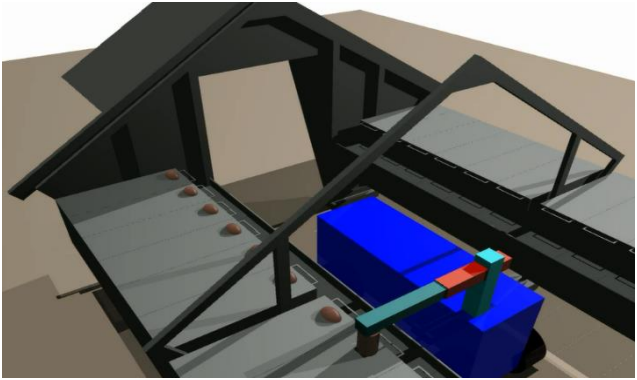

*Figure 1: 3D-visualisation of the co-simulated robot dispensing fodder.*

Feeding robots for animal husbandry have previously been developed and documented in literature. In [7] they utilize a static feeding system in combination with a RFID reader to dispense food to cows with an attached RFID tag. Outdoor piglets feeding are performed in [8] using a mobile feeding platform. The pig-feeding robot is utilized to influence the behavioural pattern and manure outlet from the piglets by daily changing the feeding position with the field.

## Martials and Methods

Crescendo combines Discrete Event (DE) modelling of a digital controller and Continuous Time (CT) modelling of the plant/environment for co-simulation. The Overture tool and VDM formalism models the DE controller and 20-sim the CT components. The Crescendo co-simulation engine coordinates simulation between 20-sim and VDM tool by implementing a protocol for time-step synchronisation between the tools. Crescendo binds the domain models together using the Crescendo contract and is responsible for exchange of information between the tools.

This model analysis is for model of a robot for dispensing animal food along row of cages at predetermined locations. The robot co-model is evaluated based overall system performance demands for the different system configurations. The chosen robot is a four-wheeled vehicle with front wheel steering and the rear wheels driving using a differential. The robot receives sensory input from vision; Radio Frequency Identification (RFID) tag reader and rotary-encoders on back wheels and front wheel kingpins. Actuators control the vehicle steering, driving and feeding system based on the sensory input. A feeder arm system mounted onto the robot is used to dispense the food onto the cages at the predetermined locations. RFID tags are placed along the animal cage rows, providing fixed reference locations. Fused sensory data are utilised to determine current location and enable the robot to perform actions in the environment.

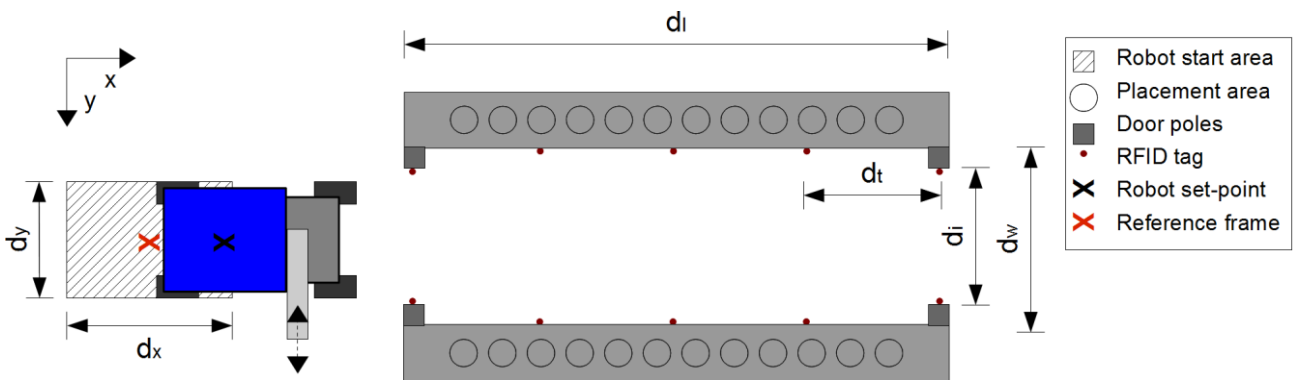

*Figure 2: Sketch of the robot vehicle and the feeding area.*

System performance demands defines what the robot must achieve to be perceived as a viable solution. The project demands to the system performance are:

- Maximum vehicle speed of 0.25 m/s
- Feeding with a precision of ± 0.05m inside the placement areas.
- No collisions with surroundings (see Figure 2)

One should notice that the performance demands are non-domain specific and focuses on the overall response of the robot in action.

Crescendo binds the domain models together using the Crescendo contract and is responsible for exchange of information between the tools. The contract contains the parameters and variables CT and DE developer's needs to be aware of when developing a combined model. Variables operated by the CT side are defined by the **monitored** keyword and variables controlled by the DE side by the **controlled** keyword. Compared to reality these variables are abstractions that only contains information for current co-model development. The controlled variables are the input to robot movement and feeding-arm. The monitored variables represent the sensory inputs to the DE side from vision; RFID; IMU; encoder and feeder-arm feedback.

The initial position is of interest in this instance since a human operator may place the robot at the starting area with resulting inaccuracies. Co-simulation evaluation is based on the robot feeding output in the surrounding environment relation to set demands. Processing feeding output requires information of the placement of each food dispensing in the operational environment, making it a post-processing process after each co-simulation.

### Results

In Figure 3 the four candidate solutions are illustrated with different arm and feeding systems.

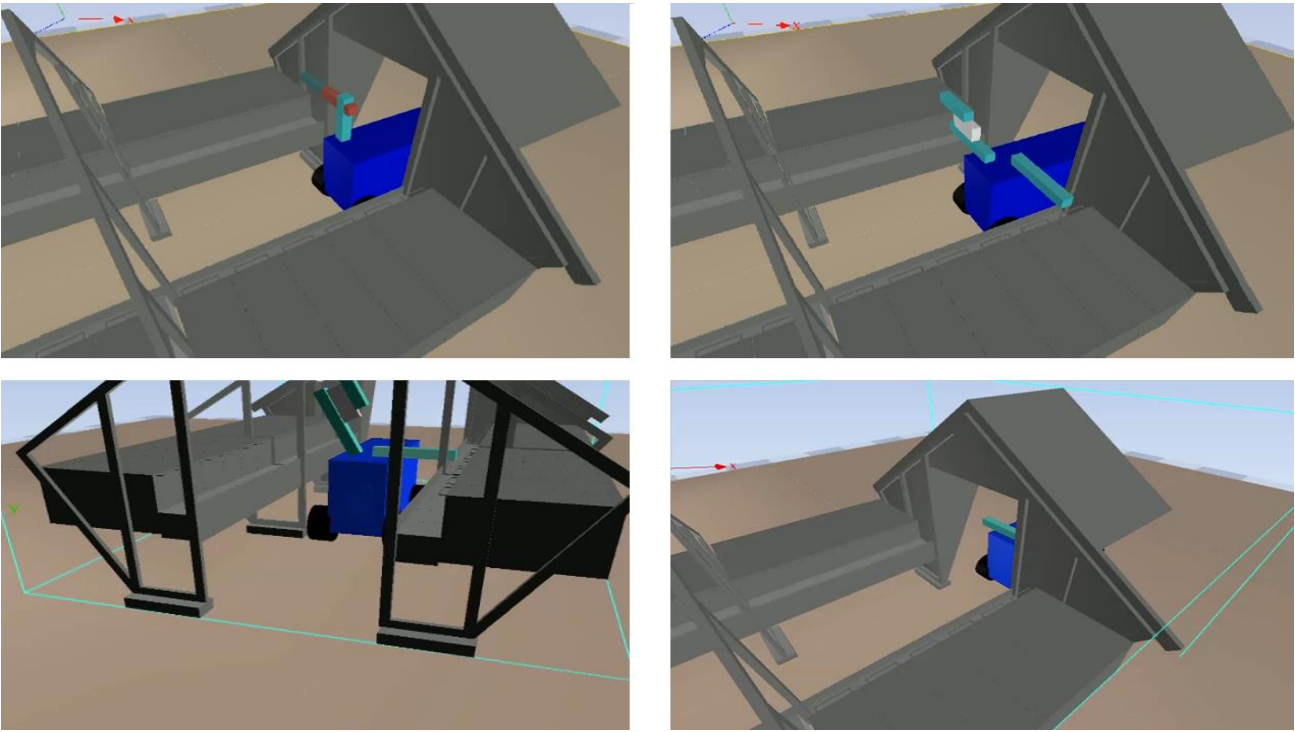

*Figure 3: The candidate solutions to the feeding system that are part of the DSE.*

We consider both solutions with single and double sided feeding output. Feeding in both sides would double the output placement of fodder at the same vehicle speed. By shifting the arms half a cage length one could use the same pump system for both side and still be able to output individual fodder amounts. The operation of arm could also be performed be by moving the arm using either rotation or translation of each joint. Each solution was modelled to conform to defined system performance demands and evaluated based on co-simulation response.

It was chosen to use a single sided solution with translatory joints for feed-arm operation. Since feeding generally is performed at the same height for each row of cages, a translatory joint was deemed a more robust solution. Perform movement of the arms joint positions is only by the DE controller, when the robot is moving in and out of a row of feeding cages.

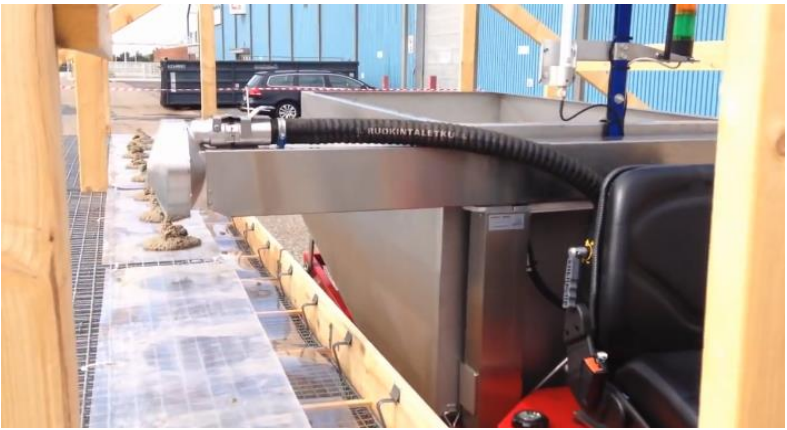

*Figure 4: Deployment co-model to an actual system*

The reason to only deploying a single sided system in the real world, was to keep the first version of the actual more simple in terms of components. Based on the co-simulation is was determined that a double sided system could be added later on, since the controller code for feeding system would be the same for the other arm.

## Discussion
The results provides an overview of the candidate system configurations. Developers used the candidate overview to select the configurations to test out on the actual platform. The overview will not grantee optimal solutions, but is an assisting tool to analyse multiple considered solutions. Factors like material, development, implementation and maintenance cost can influence the selection of a candidate configuration.

## Concluding remarks
Developing a robotic system to conform to overall system demands is essential. In this article, we have described the concept of co-modelling and co-simulation as a robotic design approach. We have shown how co-simulation using DSE can provide an overview of cross-disciplinary design candidates in robotic development. The model of the feeding robot combines the modelling in Overture (VDM) and 20-sim to a complete co-model allowing developers to utilize tool specific to their discipline. We deem that co-modelling and co-simulation combined with DSE can be utilised as an early stage development approach to analyse and compare design candidates from different domains.

## Acknowledgement

Financial support given by the Danish Ministry of Food, Agriculture and Fisheries is gratefully acknowledged. We would like to thank the Grassbots project under the EU FP7 program for partial funding. We acknowledge partial support from the EU FP7 DESTECS project on co-simulation.